\documentclass[conference]{IEEEtran}
\IEEEoverridecommandlockouts
\usepackage{cite}
\usepackage{amsmath,amssymb,amsfonts}
\usepackage{algorithmic}
\usepackage{graphicx}
\usepackage{textcomp}
\usepackage{xcolor}
\def\BibTeX{{\rm B\kern-.05em{\sc i\kern-.025em b}\kern-.08em
		T\kern-.1667em\lower.7ex\hbox{E}\kern-.125emX}}

\usepackage{times}

\usepackage{url}
\usepackage{pstricks-add}

\newcommand{\mybox}[1]{\pscirclebox[linecolor=gray,shadowcolor=lightgray]{\ensuremath{#1}}}
\newcommand{\myx}[1]{\psframebox[linecolor=gray,shadowcolor=lightgray]{\ensuremath{#1}}}

\begin{document}
	\title{Dynamic Prediction of ICU Mortality Risk\\Using Domain Adaptation}
	
	
	\author{
	\IEEEauthorblockN{Tiago Alves, Alberto Laender, Adriano Veloso, Nivio Ziviani}
	\IEEEauthorblockA{\textit{Computer Science Department at UFMG \& Kunumi} \\
		Belo Horizonte, Brazil \\
		\{nivio,tiago\}@kunumi.com}
		\{adrianov,laender,nivio\}@dcc.ufmg.br}
	
	\maketitle
	\thispagestyle{empty}
	
	\begin{abstract}
Early recognition of risky trajectories during an Intensive Care Unit (ICU) stay is one of the key steps towards improving patient survival. Learning trajectories from physiological signals continuously measured during an ICU stay requires learning time-series features that are robust and discriminative across diverse patient populations. Patients within different ICU populations (referred here as domains) vary by age, conditions and interventions. Thus, mortality prediction models using patient data from a particular ICU population may perform suboptimally in other populations because the features used to train such models have different distributions across the groups. In this paper, we explore domain adaptation strategies in order to learn mortality prediction models that extract and transfer complex temporal features from multivariate time-series ICU data. Features are extracted in a way that the state of the patient in a certain time depends on the previous state. This enables dynamic predictions and creates a mortality risk space that describes the risk of a patient at a particular time. Experiments based on cross-ICU populations reveals that our model outperforms all considered baselines. Gains in terms of AUC range from 4\% to 8\% for early predictions when compared with a recent state-of-the-art representative for ICU mortality prediction. In particular, models for the Cardiac ICU population achieve AUC numbers as high as 0.88, showing excellent clinical utility for early mortality prediction. Finally, we present an explanation of factors contributing to the possible ICU outcomes, so that our models can be used to complement clinical reasoning.
	\end{abstract}

	\section{Introduction}

Data from patients in the Intensive Care Unit (ICU) are extensive, complex, and often produced at a rate far greater than intensivists can absorb. Monitoring ICU patients is becoming increasingly complicated, and systems that learn from ICU data to alert clinicians to the current and future risks of a patient are playing a significant role in the decision making process~\cite{mcneill}. One of the main barriers in the deployment of these learning systems is the lack of generalisation of results as the learning effectiveness achieved in controlled environments often drops when the models are tested with different patient populations and conditions~\cite{lifetime}.

In this paper, we explore domain adaptation approaches to improve the accuracy of systems evaluated with mismatched training and testing conditions. We propose deep models that extract the domain-shared and the domain-specific latent features from ICU domains or patient sub-populations. Each domain corresponds to a different ICU type, such as cardiac, coronary, medical and surgical ICUs. This enables us to learn multiple models that are specific to each ICU domain, improving prediction accuracy over diverse patient populations. For this, we employ transference approaches that differ in terms of the choice of which layers to freeze or tune~\cite{ijcnn2}.

Our proposed models combine convolutional and recurrent components. While this combination has been investigated in prior work other than mortality prediction~\cite{wang}, here we capture local physiological interactions (e.g., heart rate, creatinine, systolic blood pressure) at the lower level using a Convolutional Neural Network (CNN) and extracts the long range dependencies based on convolved physiological signals at the higher level using a Long Short-Term Memory network (LSTM). Thus, our models exploit temporal information within vital signals and laboratorial findings to dynamically predict patient outcomes, i.e., the CNN component extracts features of varying abstract levels and the LSTM component ingests a sequence of these features to generate dynamic predictions for patient mortality.

As a consequence, the learned representations along with the predictions for a specific patient during the ICU stay form the corresponding patient trajectory and, thus, a mortality risk space can be obtained from a set of past patient trajectories. The fundamental benefit of analyzing future patient trajectories in the mortality risk space is the focus on dynamics, emphasizing the proximity to risky regions and the speed in which the patient condition changes. Thus, the mortality risk space enables clinicians to track risky trends and to gain insight into their treatment decisions.

\subsection{Contributions and Findings.} 
In this paper we elucidate the extent to which ICU mortality prediction may benefit from ICU domain adaptation. Thus, our main contributions are:
\begin{itemize}
	\item We present a combination of convolutional and recurrent architecture that offers a complementary temporal perspective of the patient condition. As a result, predictions based on information that is continuously collected over time can be dynamically updated as soon as new information beco\-mes available~\cite{dami}. Further, we employ Shapley additive explanations~\cite{shap} over the ICU stay in order to provide interpretable real-time predictions to help physicians prevent risky trajectories in the ICU and to complement clinical reasoning~\cite{jidm}.
	\item We show that patients within different ICU domains form sub-populations with different marginal distributions over their feature spaces. Therefore, we propose to learn specific models for different ICU domains that are trained using different feature transference approaches, instead of learning a single model for different ICU domains. We show that the effectiveness of different feature transference approaches varies greatly depending on the factors that define the target domain.
	\item We conducted rigorous experiments using the PhysioNet 2012 dataset~\cite{silva}, which comprises four different ICU domains. We show that multi-domain ICU data used for adaptation can significantly improve the effectiveness of the final model. Gains in terms of AUC range from 4\% to 8\% for early predictions, i.e., predictions based on data acquired during the first $5-20$ hours after admission.
	\item We show that the patient representations along with the predictions provided by our models are meaningful in the sense that they form trajectories in a mortality risk space. Dynamics within this space can be very discriminative, enabling clinicians to track risky trends and to gain more insight into their treatment decisions.
\end{itemize}


\section{Background Research}

Research on predicting ICU mortality is of great academic interest in medicine~\cite{cai} and in clinical machine learning~\cite{decision,luo}. A number of researchers have investigated how to correlate ICU data with patient outcomes. In one of the first studies~\cite{ai}, the authors identify parameters in patient data that correlate with outcomes. In what follows, we discuss previous work in contrast with ours.

\subsection{Mortality Prediction}

The PhysioNet ICU Mortality Challenge 2012~\cite{silva} provided benchmark data that incorporate evolving clinical data for ICU mortality prediction. As~\cite{revisited} reported, this benchmark data fostered the development of new approaches, leading to up to 170\% improvement over traditional risk scoring systems that do not incorporate such clinical data currently used in ICUs~\cite{saps}. In what follows, we discuss previous work in contrast with ours.

Most of the current work uses the PhysioNet ICU Mortality Challenge 2012 data. The most effective approaches are based on learning discriminative classifiers for specific sub-populations.
Authors in~\cite{physionet} proposed a robust SVM classifier, while authors in~\cite{bera} proposed a logistic regression classifier. Authors in~\cite{vairavan} also employed logistic regression classifiers, but coupled them with Hidden Markov Models in order to model time-series data. Shallow neural networks were evaluated in~\cite{xia}, while a tree-based Bayesian ensemble classifier was evaluated in~\cite{bayes}. Authors in~\cite{krajnak} employed fuzzy rule-based systems, and authors in~\cite{mcmillan} proposed an approach that identifies and integrates information in motifs that are statistically over- or under-represented in ICU time series of patients. Authors in~\cite{lipton1} used LSTMs to improve the classification of diagnoses. 

More recently, authors in~\cite{hyun} proposed a Markov model that accumulates mortality probabilities. Likewise, authors in~\cite{time-series} proposed an approach that models the  mortality probability as a latent state that evolves over time. Authors in~\cite{kdd2} proposed an approach to address the problem of small data using transfer learning in the context of developing risk models for cardiac surgeries. They explored ways to build surgery-specific and hospital-specific models using information from other kinds of surgeries and other hospitals. Their approach is based on weighting examples according to their similarity to the target task training examples. The three aforementioned works are considered as baselines and compared with our approach.

Following~\cite{kdd2}, in this work we use feature transference, but in a quite different way, as follows: (i) instead of applying instance weighting, we employed a deep model that transfers domain-shared features; (ii) we studied a broader scenario that includes diverse ICU domains; and (iii) our models employ both spatial and temporal feature extraction, being able to predict patient outcomes dynamically.

\subsection{ICU Domains and Sub-Populations}

Imbalanced data~\cite{imbalance}, sub-populations of patients with different marginal distributions over their feature spaces~\cite{nori}, and sparse data acquired from heterogeneous sources~\cite{szolovits2,het} are issues that pose significant challenges for ICU mortality prediction. 

Authors in~\cite{gong} discussed problems due to the lack of consistency in how semantically equivalent information is encoded in different ICU databases. Authors in~\cite{imbalance} discussed the problem of imbalanced ICU data, which occurs when one of the possible patient outcomes is significantly under-represented in the data. Further, since features are often imbalanced, some ICU domains have a significantly larger number of observations than others (e.g., respiratory failure in adults vs. children). In a recent work, authors in~\cite{icde} proposed a mortality study based on the notion of burstiness, where high values of burstiness in time-series ICU data may relate to possible complications in the patient’s medical condition and hence provide indications on the mortality. Authors in~\cite{variational} employed a  variational recurrent neural network in order to capture temporal latent dependencies of multivariate time-series data.

While most studies on mortality prediction for ICU patients have assumed that one common risk model could be developed and applied to all the patients, authors in~\cite{nori} advocated that this might fail to capture the diversity of ICU patients. As shown in~\cite{lifetime}, models built using patient data from particular age groups perform poorly on other age groups because the features used to train the models have different distributions across the groups. 

\subsection{Our Work}

None of the aforementioned approaches attempted to perform ICU domain adaptation, which is a core focus of our work. There is often a mismatch between different ICU domains or patient sub-populations, and domain adaptation seems to be a natural solution for learning more robust models, as different ICU domains share features that exhibit different distributions. While data in different ICU domains may vary, there are potentially shared or local invariant features that shape patients in different ICU domains.

Other focus of our work is to capture spatial and temporal features from time-series ICU data. Features are captured in a way that the state of the patient in a certain time depends on the previous state. This forms a risk space, and trajectories in this space allow to easily describe the state of the patient at a particular time, helping intensivists to estimate the patient progress from the current patient state.

\section{Methods}

The task of predicting patient outcomes over time from ICU data is defined as follows. We have as input the {\em training set}, which consists of a sequence of observations of the form $<A_t,o_t>$, where $A_t$ is a vector of values corresponding to physiological parameters associated with a patient at time $t$, and $o_t$ is the outcome for the patient at time $t$ (i.e., whether or not the patient survived the hospitalization, replicated for each time $t$). The training set is used to construct a model that relates features within the sequence of observations to the patient outcome. The {\em test set} consists of a sequence of observations $<A_t,?>$ for which only the physiological parameters for the patient at time $t$ are available, while the corresponding patient outcome is unknown. The model learned from the training set is used to predict the outcome for patients in the test set. The task of dynamically predicting patients outcomes in the ICU has two important requirements:
\begin{itemize}
	\item It is a domain-specific problem, i.e., a prediction model lear\-ned from a sub-population (or ICU domain) is likely to fail when tested against data from other population~\cite{variational}. Feature transferability is thus an appealing way to provide robustness to the prediction models~\cite{mm17,ijcnn}.
	\item It is a time-sensitive problem, i.e., accurately predicting patient outcomes as early as possible may lead to earlier diagnosis and more effective therapies.
\end{itemize}

Next we present our model, which is built from multi-domain ICU time-series data and is designed to provide dynamically updated estimates of patient mortality.

\subsection{Network Architecture}

\begin{figure}[!t]
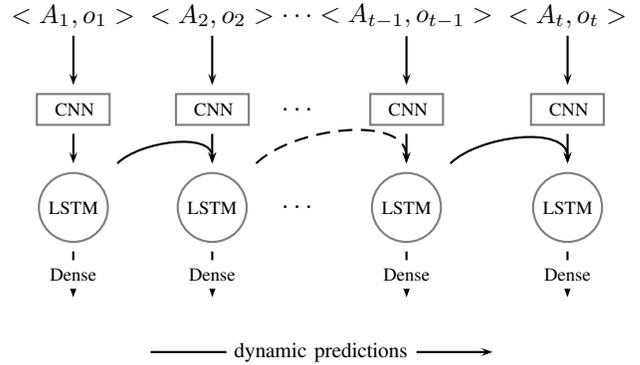

\centering
$\psmatrix[mnode=r,colsep=0.1,rowsep=0.2]
&[name=a1]\scriptsize{<A_1,o_1>} && [name=c1] \scriptsize{<A_2,o_2>} &\ldots& [name=e1] \scriptsize{<A_{t-1},o_{t-1}>} && [name=g1] \scriptsize{<A_t,o_t>}\\[0pt]
& & & & & & &\\[0pt]
&[name=a2] \myx{\mbox{~\scriptsize{CNN}~}} & & [name=c2] \myx{\mbox{~\scriptsize{CNN}~}} & \ldots & [name=e2] \myx{\mbox{~\scriptsize{CNN}~}}& & [name=g2] \myx{\mbox{~\scriptsize{CNN}~}}\\[0pt]
& & & & & & &\\[0pt]
&[name=a3] \mybox{\mbox{\scriptsize{LSTM}}} & & [name=c3] \mybox{\mbox{\scriptsize{LSTM}}} & [name=x] \ldots & [name=e3] \mybox{\mbox{\scriptsize{LSTM}}}& & [name=g3] \mybox{\mbox{\scriptsize{LSTM}}}\\[0pt]
& & & & & & &\\[0pt]
& [name=a4] & & [name=c4] & & [name=e4] & & [name=g4]\\[0pt]
 && [name=a] && && [name=b]
\endpsmatrix $
\psset{nodesep=3pt,arrows=->}
\ncline{->}{a}{b}
\ncput*{\footnotesize{dynamic predictions}}
\ncline{->}{a1}{a2}
\ncline{->}{a2}{a3}
\ncline{->}{a3}{a4}
\ncput*{\scriptsize{Dense}}
\ncline{->}{c1}{c2}
\ncline{->}{c2}{c3}
\ncline{->}{c3}{c4}
\ncput*{\scriptsize{Dense}}
\ncline{->}{e1}{e2}
\ncline{->}{e2}{e3}
\ncline{->}{e3}{e4}
\ncput*{\scriptsize{Dense}}
\ncline{->}{g1}{g2}
\ncline{->}{g2}{g3}
\ncline{->}{g3}{g4}
\ncput*{\scriptsize{Dense}}
\nccurve[angleA=45,angleB=90]{a3}{c3}
\nccurve[angleA=45,angleB=90,linestyle=dashed]{c3}{e3}
\nccurve[angleA=45,angleB=90]{e3}{g3}
\caption{Architecture for predicting outcomes over time. Each convolutional (CNN) layer is followed by a LSTM layer and different feature transference approaches are designed using this architecture.}
\label{fig:arch}
\end{figure}

Here, we introduce our deep model, referred to as CNN$-$ LSTM, which is composed of both convolutional and recurrent layers, as shown in~Figure~\ref{fig:arch}. Convolutional and recurrent components offer a complementary perspective of the patient condition, as follows: the convolutional layer encodes temporal physiological information locally, while the recurrent layer is designed to capture long range information and forget unimportant local information. 

Specifically, our model employs one-dimensional CNN layers~\cite{nips} followed by max-pooling layers, thus extracting correlations between physiological parameters measured in consecutive time periods. For instance, it may find that if both temperature and heart rate are increasing, the odds of survival decrease. In a complementary way, the recurrent layer (LSTM) learns how changes in observations for a patient affect the corresponding outcome. Intuitively, the recurrent layer captures temporal dependencies, enabling the estimation of patient progress from the current patient state. For instance, if the heart rate was low at the beginning of the stay and then becomes very high, then the odds of survival decrease. Finally, a dense layer takes the output of the recurrent layer and predicts the patient outcome.


In summary, our model works by passing each observation through a feature extractor and then the sequence model captures how the extracted features are associated with patient outcomes over time.
Also, dropout operation is performed after each layer of the network.

As not all the descriptors and time-series were available for all records, we had to deal with the problem of missing values. If one variable (either a descriptor or a time-series) was never recorded for a given patient, we used the approach called “imputation” and replaced its features with value zero after normalization. Because of the normalization step, this approximately corresponds to replacing the missing raw variable with a measure of central tendency, which corresponds to the arithmetic mean for Gaussian-distributed variables and to the geometric mean for log-normal ones. In some cases, the time-series measurement was taken only in the first 24 hours or only during the next 24 hours. In this case, replacing with zero all the features related to the period with missing measurements could possibly create a non-existing improvement or deterioration trend. Instead, we duplicate the values from the available period, assuming stationarity conditions as default in absence of further measurements.

\begin{table*} [ht!]
\caption{Average patient physiological data. Mean, first and third quartiles within each physiological parameter. Mortality rate is concentrated in the Medical ICU (49.6\% of all the deaths).}
\centering
\small
  \begin{tabular}{lllll} \hline
 & Cardiac & Coronary & Medical & Surgical\\\hline
N      &  874 &  577 &  1,481 &  1,067\\
Age    &  67.91 (56$-$79) & 69.22 (59$-$81) &  62.83 (51$-$78) &  60.50 (48$-$76)\\
Male   &  530 (60.6\%) &  333 (57.7\%) &  753 (50.8\%) &  630 (59.0\%)\\
Mortality Rate & 4.9\% (7.8\%) & 14.0\% (14.6\%) & 18.6\% (49.6\%) & 14.5\% (28.0\%)\\
\hline\\
Albumin (g/dL)  &  2.92 (2.4$-$3.5)  &  3.31 (2.9$-$3.6)  &  2.92 (2.5$-$3.3)  &  2.99 (2.5$-$3.5) \\
Alkaline phosphatase (IU/L)  &  74.93 (46$-$83)  &  92.44 (59$-$102)  &  126.15 (64$-$138)  &  91.43 (52$-$96) \\
Alanine transaminase (IU/L)  &  28.70 (18$-$45)  &  68.14 (19$-$78)  &  45.17 (16$-$61)  &  72.11 (17$-$84) \\
Aspartate transaminase (IU/L)  &  37.19 (28$-$56)  &  32.41 (26$-$55)  &  42.14 (24$-$57)  &  34.90 (24$-$53) \\
Bilirubin (mg/dL)  &  1.01 (0.4$-$1.1)  &  0.87 (0.4$-$0.9)  &  2.44 (0.4$-$1.6)  &  1.85 (0.5$-$1.5) \\
Cholesterol (mg/dL)  &  150.14 (114$-$174)  &  163.59 (134$-$189)  &  141.04 (111$-$169)  &  157.87 (122$-$184) \\
Creatinine (mg/dL)  &  1.04 (0.7$-$1.1)  &  1.58 (0.8$-$1.6)  &  1.64 (0.7$-$1.7)  &  1.12 (0.7$-$1.1) \\
Invasive diast. press. (mmHg)  &  58.85 (51$-$66)  &  62.65 (53$-$74)  &  54.97 (48$-$70)  &  59.65 (52$-$72) \\
Fractional inspired O2  &  0.91 (1.0$-$1.0)  &  0.82 (0.5$-$1.0)  &  0.72 (0.5$-$1.0)  &  0.72 (0.5$-$1.0) \\
Serum glucose (mg/dL)  &  129.28 (103$-$145)  &  165.74 (114$-$191)  &  155.02 (104$-$175)  &  148.85 (114$-$167) \\
Serum bicarbonate (mmol/L)  &  23.41 (22$-$25)  &  23.31 (21$-$26)  &  22.74 (19$-$26)  &  23.44 (21$-$26) \\
Hematocrit (\%)  &  29.32 (25.3$-$32.8)  &  34.48 (30.7$-$37.8)  &  31.82 (27.9$-$36)  &  33.01 (29.1$-$36.8) \\
Heart rate (bpm)  &  85.43 (79$-$91)  &  84.32 (69$-$97)  &  95.61 (80$-$110)  &  87.83 (74$-$100) \\
Serum potassium (mEq/L)  &  4.49 (4$-$4.7)  &  4.28 (3.8$-$4.5)  &  4.19 (3.6$-$4.5)  &  4.07 (3.6$-$4.3) \\
Lactate (mmol/L)  &  2.76 (1.5$-$3.3)  &  2.76 (1.4$-$3)  &  2.58 (1.3$-$2.8)  &  2.65 (1.3$-$3.1) \\
Invasive mean press. (mmHg)  &  78.86 (69$-$86)  &  86.14 (73$-$99)  &  86.58 (68$-$96)  &  87.13 (73$-$98) \\
Serum sodium (mEq/L)  &  138.42 (136$-$140)  &  137.82 (135$-$140)  &  138.96 (136$-$142)  &  139.33 (137$-$142) \\
Non-invasive diast. press. (mmHg)  &  52.21 (44$-$59)  &  61.15 (49$-$72)  &  62.03 (50$-$72)  &  62.42 (52$-$73) \\
Non-invasive mean press. (mmHg)  &  71.53 (62$-$79)  &  78.93 (67$-$89)  &  80.55 (68$-$91)  &  82.78 (71$-$94) \\
Non-invasive syst. press. (mmHg)  &  110.88 (96$-$125)  &  117.46 (101$-$134)  &  121.78 (104$-$138)  &  126.72 (108$-$145) \\
Partial press. of art. CO2 (mmHg)  &  41.20 (36$-$45)  &  40.61 (35$-$45)  &  42.50 (34$-$48)  &  41.01 (35$-$45) \\
Partial press. of art. O2 (mmHg)  &  295.46 (218$-$387)  &  181.58 (89$-$248)  &  147.68 (78$-$185)  &  188.24 (101$-$250) \\
Arterial pH (0-14) &  7.39 (7.35$-$7.44)  &  7.84 (7.31$-$7.43)  &  7.44 (7.3$-$7.42)  &  7.46 (7.32$-$7.43) \\
Platelets (cells/nL)  &  170.36 (117$-$208)  &  241.44 (181$-$283)  &  230.89 (143$-$287)  &  219.19 (150$-$268) \\
Respiration rate (bpm)  &  17.55 (14$-$20)  &  19.74 (16$-$23)  &  21.10 (17$-$24)  &  18.95 (16$-$21) \\
Invasive systolic press. (mmHg)  &  117.16 (105$-$127)  &  117.65 (100$-$139)  &  107.45 (95$-$137)  &  123.33 (108$-$148) \\
Temperature ($^o$C)  &  35.57 (35.5$-$36.6)  &  36.38 (36$-$37.1)  &  36.77 (36.2$-$37.4)  &  36.51 (36.1$-$37.4) \\
Troponin-I ($\mu$g/L)  &  6.77 (0.8$-$10.1)  &  10.05 (0.8$-$12.4)  &  5.59 (0.8$-$7)  &  7.02 (0.4$-$6.7) \\
Troponin-T ($\mu$g/L)  &  1.51 (0.04$-$0.59)  &  2.78 (0.17$-$2.8)  &  0.33 (0.04$-$0.25)  &  0.22 (0.03$-$0.14) \\
Urine output (mL)  &  497.92 (120$-$615)  &  365.62 (100$-$500)  &  255.39 (70$-$325)  &  389.29 (100$-$500) \\
White blood cell (cells/nL)  &  12.98 (9.2$-$15.5)  &  12.31 (8.5$-$14.3)  &  13.33 (7.8$-$17)  &  12.37 (8.4$-$15.1)\\
\hline
\end{tabular}
\label{tab:physionet}
\end{table*}

\subsection{Feature Transferability}

Our goal is to train multi-domain models to predict patient outcomes over time, which is based on patient observations from multiple ICU domains. Although patients from a given ICU domain may be better represented by domain-specific features, there still exist some common features that permeate all other ICU domains. 

The main intuition that we exploit for feature transferability is that the features must eventually transition from general to specific along our model and, accordingly to ~\cite{bengio2}, feature transferability drops significantly in higher layers with increasing domain discrepancy. In other words, the features computed in higher layers must depend strongly on a specific domain and prediction effectiveness suffers if this domain is discrepant from the target domain. Our proposal is to initialize the model with pretrained weights of source ICU domains, which are 
then fine-tuned with data from the target ICU domain. Since we are dealing with many domains simultaneously, we tested different transference approaches, which are detailed as follows:

\begin{description}
	\item [A1:] No layer is kept frozen during fine-tuning, i.e., errors are back-propagated through the entire network during fine-tuning.
	\item [A2:] Only the convolutional layer is kept frozen during fine-tuning.
	\item [A3:] Convolutional and LSTM layers are kept frozen during fine-tuning, i.e., errors are back-propagated only thought the fully-connected layers during fine-tuning.
	\item [A4:] Only the convolutional layer is kept frozen during fine-tuning and other layers have their weights randomly initialized for fine-tuning.
	\item [A5:] Convolutional and LSTM layers are kept frozen during fine-tuning and weights in fully-connected layers are randomly initialized for fine-tuning.
\end{description}


\section{Experiments}

In this section, we present the data we used to evaluate our multi-domain model for mortality prediction over time. Then, we discuss our evaluation procedure and report the results of our multi-domain model. In particular, our experiments aim to answer the following research questions:

\begin{description}
	\item [Q1:] Does domain adaptation improve mortality prediction? Do models that are specific to each ICU domain improve the state-of-the-art for mortality prediction?
	\item [Q2:] Which feature transference approach is more appropriate to each ICU domain?
	\item [Q3:] How effective and accurate are dynamic predictions?
	\item [Q4:] How meaningful are the mortality risk spaces created from patient trajectories?
\end{description}

\subsection{Data and Domains}

We use the publicly available dataset of multivariate clinical time-series of 4,000 patients from the PhysioNet 2012 challenge~\cite{silva}.The data for each patient includes age, gender, height, weight and 37 time-stamped physiological parameters measured during the first 48 hours of ICU stay. Patient outcomes, including mortality, are available. We resample the time series on an hourly basis and propagate measurements forward (or backward) in time to fill gaps. We scale each variable to fall into the $[0,1]$ interval. The source domain is composed of all ICU domains but the target one, which is used only during fine-tuning. In contrast to~\cite{imbalance}, we did not perform feature selection and, thus used the entire feature-set in all experiments.

\begin{figure}[t]
	\begin{center}
		\includegraphics[type=eps,ext=.eps,read=.eps,width=1.06\linewidth]{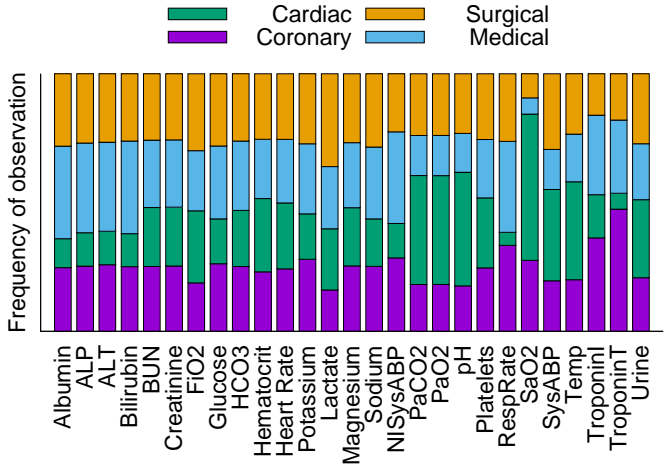}
	\end{center}
	\caption{(Color online) Relative frequency in which physiological parameters are measured in different ICU domains.}
	\label{fig:freq}
\end{figure}

Table~\ref{tab:physionet} shows the average physiological data for patients in each ICU domain. The dataset also specifies the ICU domain to which the patient has been admitted: Cardiac Surgery, Coronary Care, Medical and Surgical. Physiological data differ greatly between patients admitted to different ICU domains.
Figure~\ref{fig:freq} shows the frequency in which physiological parameters are measured within each ICU domain. Clearly, some ICU domains have a significantly larger number of observations than others (e.g., PaCO$_2$ is much more frequently measured in the Cardiac ICU, while TroponinT is much more frequently measured in the Coronary ICU).

\subsection{Baselines}

We considered the following methods in order to provide baseline comparison:

\begin{itemize}
\item Shallow classifiers: Logistic Regression (LR), Support Vector Machines (SVM: Linear Kernel, C=0.1), Random Forest (RF: depth=10, $\sqrt{n}$ random features, 200 trees). The main objective of using these baselines is to compare CNN$-$LSTM with shallow models.
\item Training on Target (TT): A CNN$-$LSTM model is trained using only the target domain data. No source domain data is used. The main objective of using this baseline is to assess the benefits of domain adaptation.
\item CNN and LSTM: A CNN and a LSTM model are trained using data from all domains. No fine-tuning is performed. The main objective of using this baseline is to assess the benefits of employing CNN and LSTM components together.
\item No tuning (NT): A CNN$-$LSTM model is trained using data from all domains. No fine-tuning is performed. ICU source is included as an input variable, so the model is aware of the source. The main objective of using this baseline is to assess the benefits of domain adaptation. 
\item \textit{Che et al., 2015}~\cite{kdd}: A deep network that uses data-driven prior-based regularization. The main objective of using this baseline is to compare our model with state-of-the-art results on the PhysioNet data.
\item \textit{Che et al., 2018}~\cite{gru}: A recent Gated Recurrent Unit network which employs a missing value imputation approach which is similar to ours. Again, the main objective of using this baseline is to compare our model with state-of-the-art results on the PhysioNet data.
\end{itemize}

\setlength{\tabcolsep}{3.4pt}
\renewcommand{\arraystretch}{1.2}
\begin{table}[ht]
	\caption{AUC numbers for shallow and deep models. Numbers in bold indicate the best models for each ICU domain.}
	\label{tab:res1}
	\begin{center}
		\small
		\begin{tabular}{lllllr}
			\hline
			Model   & Cardiac & Coronary & Medical & Surgical & Avg\\ \hline
			SVM         & 0.627 & 0.572 & 0.503 & 0.532 & 0.558\\
			LR          & 0.629 & 0.601 & 0.510 & 0.517 & 0.564\\
			RF          & 0.610 & 0.578 & 0.587 & 0.623 & 0.599\\
\hline
			TT          & 0.821 & 0.769 & 0.722 & 0.727 & 0.759\\
			LSTM        & 0.812 & 0.807 & 0.742 & 0.769 & 0.782\\
			CNN         & 0.866 & 0.802 & 0.747 & 0.812 & 0.807\\
			NT$^-$      & 0.876 & 0.833 & 0.737 & 0.801 & 0.812\\
			NT          & 0.876 & 0.837 & 0.757 & 0.812 & 0.820\\ 
\hline
			$[$Che \textit{et al., 2015}$]$ & 0.853 & 0.802 & 0.760 & 0.785 & 0.800\\
			$[$Che \textit{et al., 2018}$]$ & 0.868 & 0.824 & 0.775 & \textbf{0.823} & 0.823\\
\hline
			CNN$-$LSTM  & \textbf{0.885} & \textbf{0.848} & \textbf{0.782} & \textbf{0.827} & \textbf{0.836}\\
			\hline
		\end{tabular}
	\end{center}
	
	\setlength{\tabcolsep}{7.5pt}
	\renewcommand{\arraystretch}{1.2}
	\caption{AUC numbers for different feature transference approaches. Numbers in bold indicate the best transference approach for each target ICU domain.}
	\label{tab:res2}
	\begin{center}
		\small
		\begin{tabular}{llllll}
			\hline
			Target   & A1    & A2    & A3    & A4    & A5   \\\hline
			Cardiac  & 0.852 & \textbf{0.885} & 0.829  & 0.849 & 0.858\\
			Coronary & \textbf{0.848} & 0.812 & 0.807  & 0.793 & 0.784\\
			Medical  & 0.754 & 0.763  & \textbf{0.782} & 0.759 & 0.736\\
			Surgical & 0.822 & \textbf{0.827} & 0.808  & 0.818 & 0.788\\
			\cline{2-6}
			Overall	  & 0.819 & \textbf{0.822} & 0.806 & 0.804 & 0.791\\ 
			\hline
		\end{tabular}
	\end{center}
\end{table}
\setlength{\tabcolsep}{6.0pt}
\renewcommand{\arraystretch}{1}

\subsection{Setup}


We evaluate the effectiveness of the models using the standard Area Under the ROC Curve (AUC), as adopted by~\cite{kdd}. Like~\cite{bayes}, we used five-fold cross validation and relevant hyper-parameters were found using a validation set. Each fold is split into three distinct subsets: one for training, with 64\% of the patients, one for validation and parameter tuning, with 16\% of the patients and the final set for testing the model, with the remaining 20\% of the patients. As in other works~\cite{kdd,gru}, test set leakage was prevented by ensuring that time-series data of a specific patient are either on the training or test set, and never on both~\cite{citizen,sigir}.

For CNN$-$LSTM, learning rate was set to 0.001. We used Scaled Exponential Linear Unit~\cite{selu} as non linear activations and a dropout probability of 0.2 for every layer. The 1D-CNN components employ 64 filters, kernel size was set to 5 with stride of 1. Max pooling size was set to 4. The LSTM components employ 70 neurons on the inner cell. Training was stopped after 15 epochs with no improvement. We used ADAM~\cite{adam} in order to minimize the binary cross-entropy of the training set.

\begin{figure}[t]
\begin{center}
\includegraphics[type=eps,ext=.eps,read=.eps,width=0.74\linewidth]{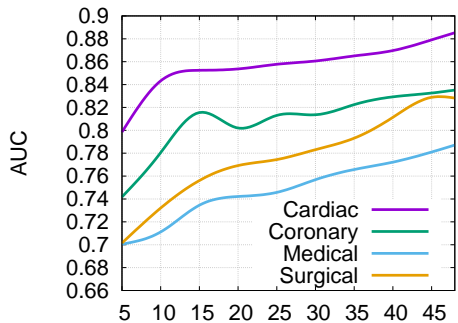}
\end{center}
\caption{(Color online) CNN$-$LSTM AUC numbers for predictions performed using information within the first $y$ hours after the patient admission ($5\le y\le48$ hours).}
\label{fig:time1}
\end{figure}

The results reported are the average of the five runs, and to ensure their relevance we assess the statistical significance of our measurements by means of a pairwise t-test~\cite{t-test} with p$-$value $\le 0.05$.
We perform a hand search for these hyper-parameters, tuning on the validation set, with early stopping. The best model was chosen according to the smallest loss on the validation set and are used to assess the overall performance of the models.

\subsection{Results and Discussion}

The first experiment is devoted to answer Q1. Table~\ref{tab:res1} shows AUC numbers for each model. We report numbers for each ICU domain, and also the macro-averaged result. Clearly, CNN$-$LSTM consistently outperforms all shallow baselines, and also~\cite{kdd}. Employing CNN and LSTM components together is beneficial, since NT is consistently superior than CNN and LSTM. Domain adaptation is beneficial for most of the domains. The only exception occurs with the Coronary domain for which performance remains statistically the same when compared with NT.
Overall, CNN$-$LSTM shows a macro-averaged AUC of 0.832.

The second experiment is concerned with Q2. Table~\ref{tab:res2} shows AUC numbers for CNN$-$LSTM models learned following the different feature transference approaches.
The best transference approach varies depending on the target ICU domain. Randomly initializing the weights for fine-tuning does not show to be the best approach, as A4 and A5 were not the best performers for any target domain. It seems that specific temporal patterns play an important role for mortality prediction in the Surgical domain, as A1 and A2 were the best approaches for this domain. For the Medical domain, A3 was the best approach, suggesting that features learned from other domains are effective. For the Cardiac and Coronary domains, A2 was the best transference approach, which indicates that specific features are important in this domain.

\begin{figure}[t]
	\begin{center}
	\includegraphics[type=eps,ext=.eps,read=.eps,width=0.74\linewidth]{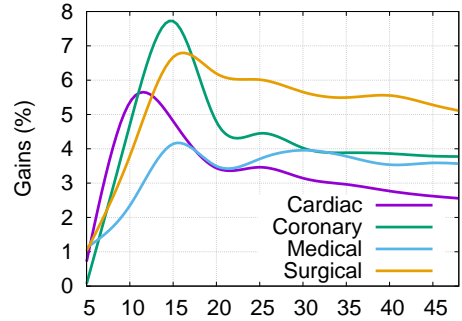}
	\end{center}
	\caption{(Color online) Gains over~\cite{gru} at different prediction times ($5\le y\le48$ hours).}
	\label{fig:time2}
\end{figure}

\begin{figure*}[h!]
	\begin{center}
		\includegraphics[type=eps,ext=.eps,read=.eps,width=0.28\linewidth]{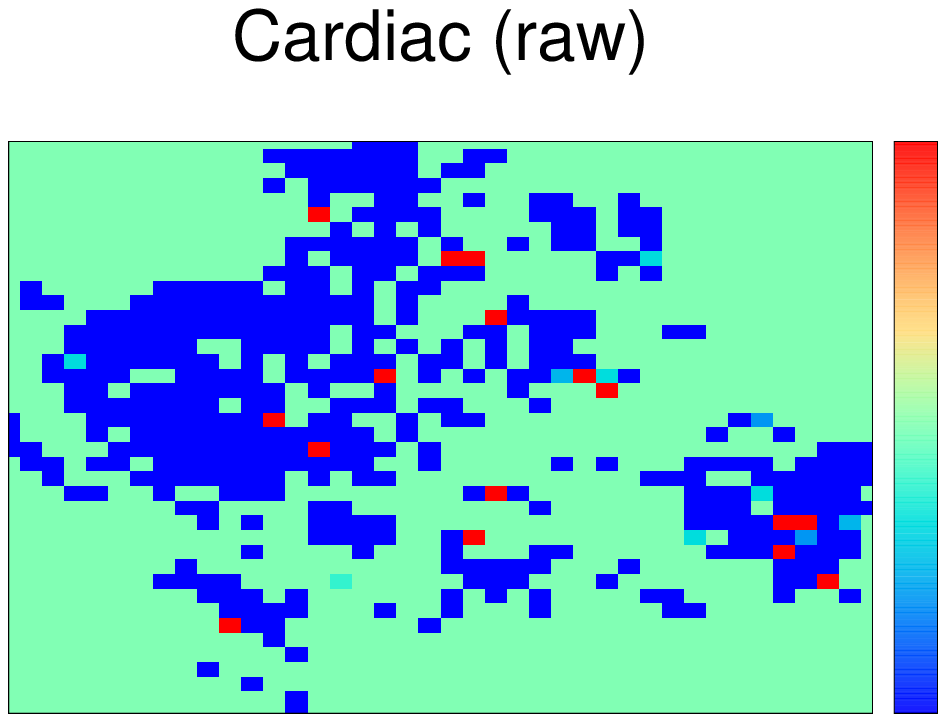}
		\hspace{-0.45in}
		\includegraphics[type=eps,ext=.eps,read=.eps,width=0.28\linewidth]{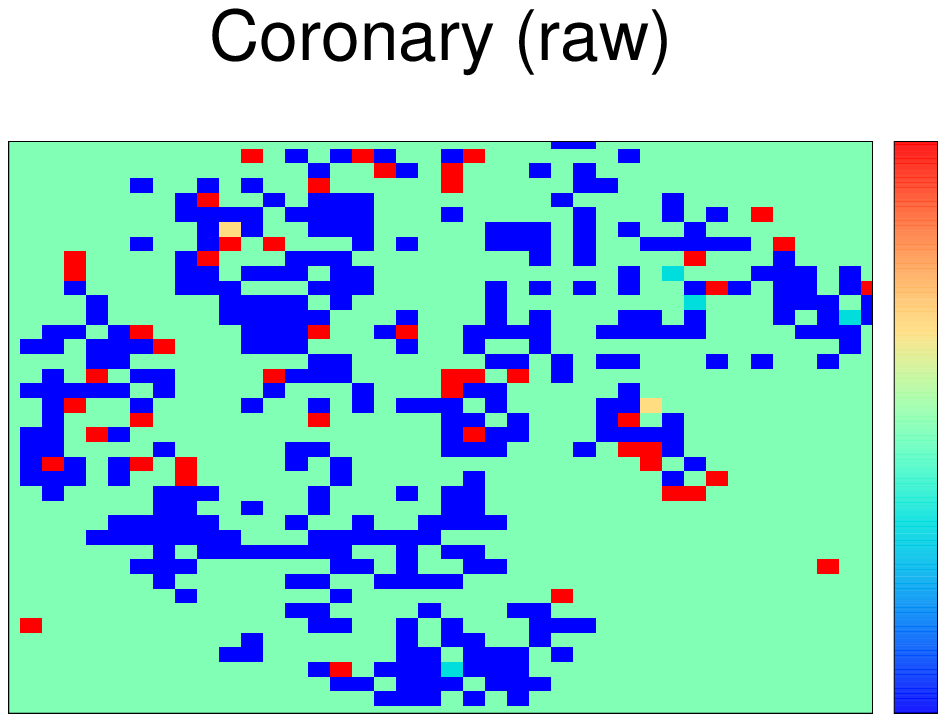}
		\hspace{-0.45in}
		\includegraphics[type=eps,ext=.eps,read=.eps,width=0.28\linewidth]{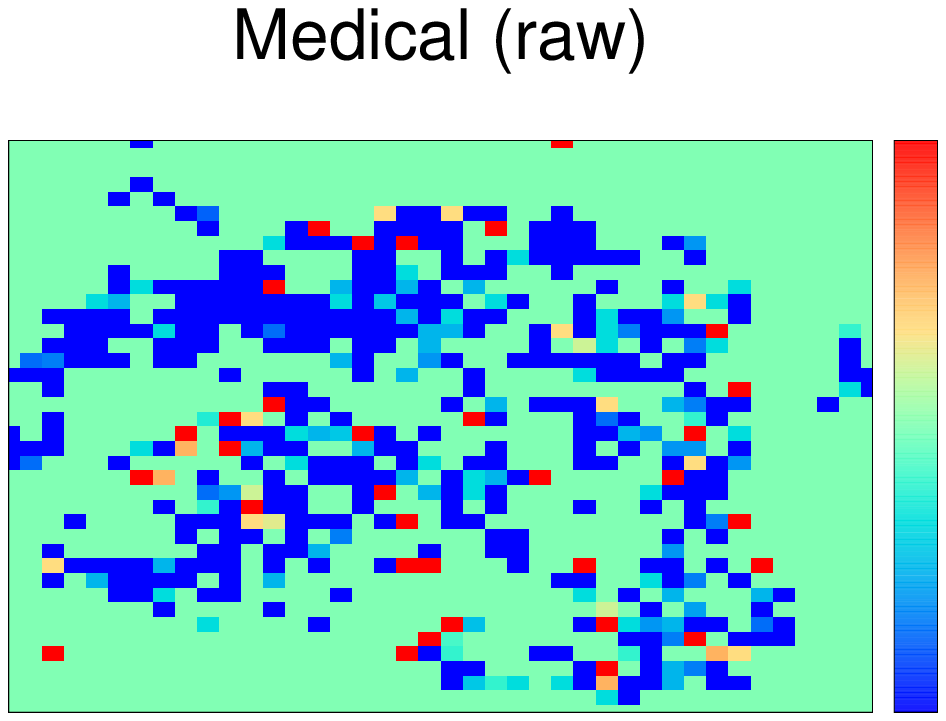}
		\hspace{-0.45in}
		\includegraphics[type=eps,ext=.eps,read=.eps,width=0.28\linewidth]{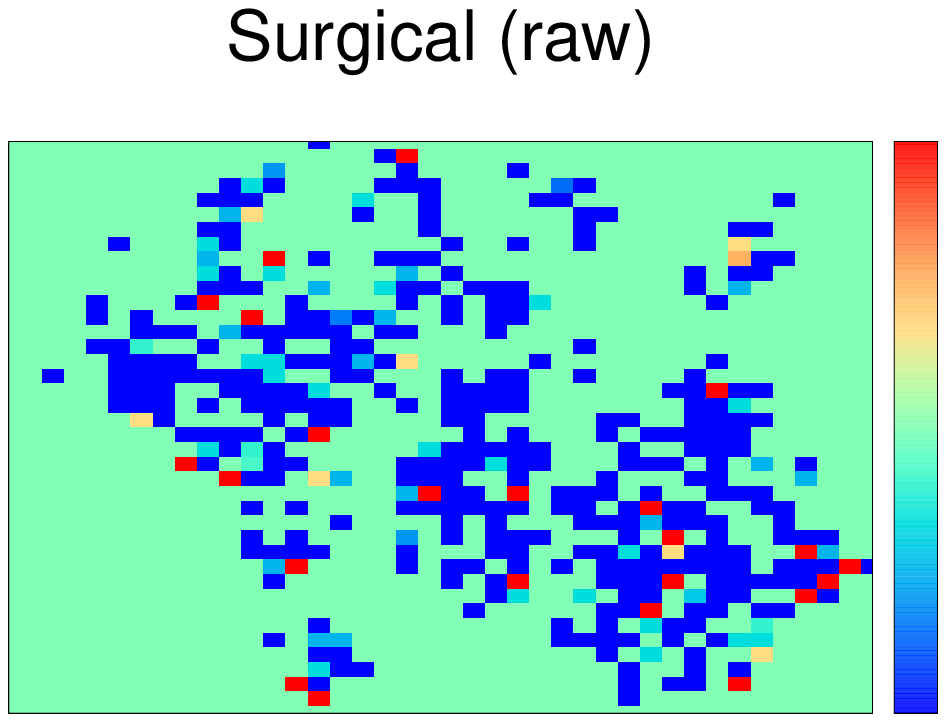}
		\includegraphics[type=eps,ext=.eps,read=.eps,width=0.28\linewidth]{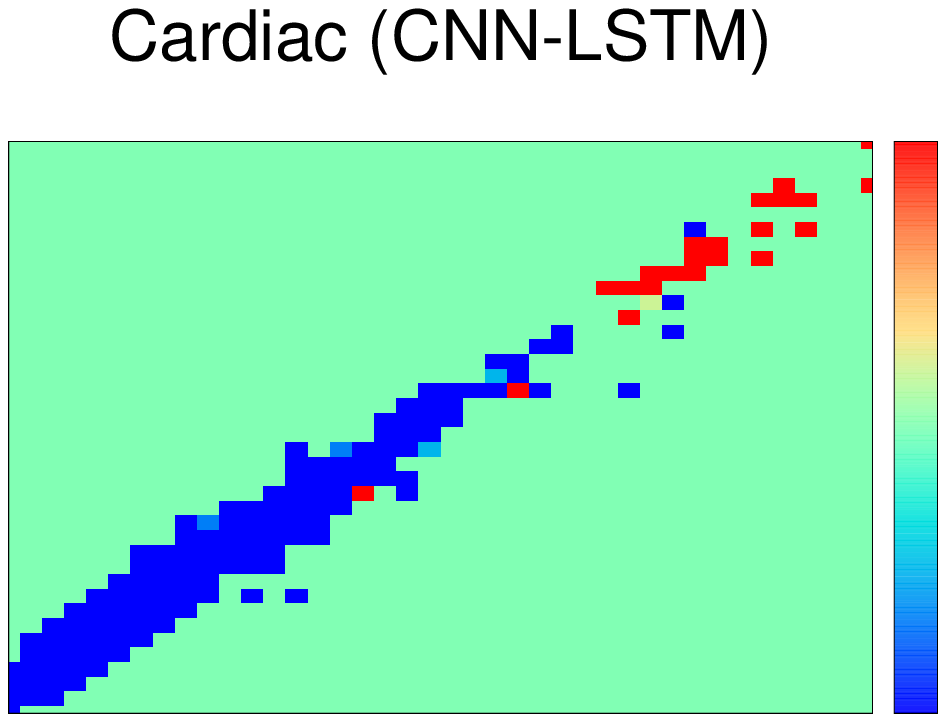}
		\hspace{-0.45in}
		\includegraphics[type=eps,ext=.eps,read=.eps,width=0.28\linewidth]{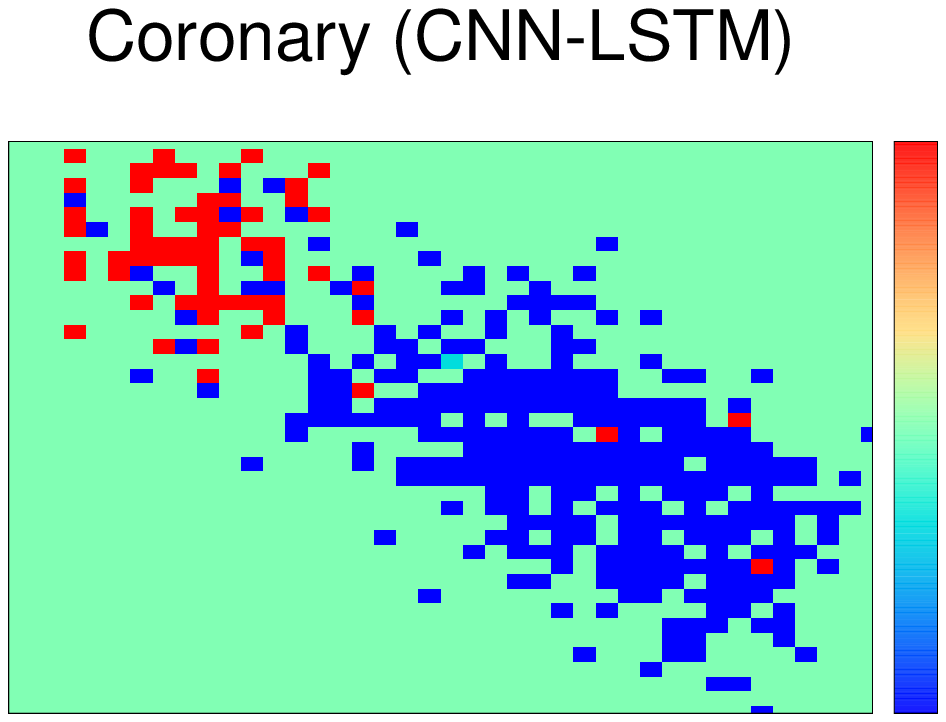}
		\hspace{-0.45in}
		\includegraphics[type=eps,ext=.eps,read=.eps,width=0.28\linewidth]{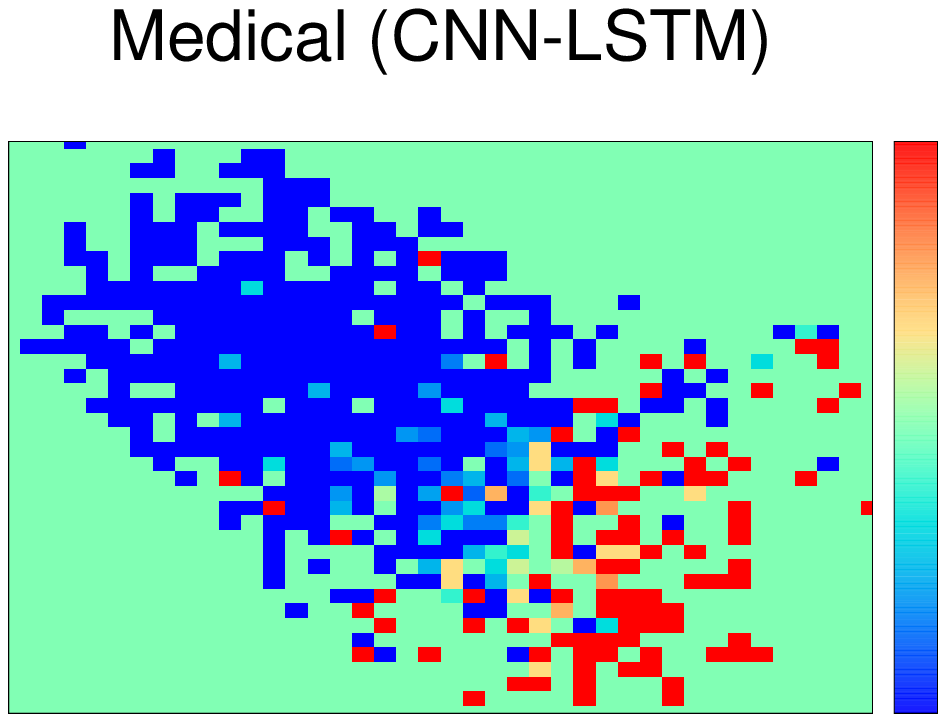}
		\hspace{-0.45in}
		\includegraphics[type=eps,ext=.eps,read=.eps,width=0.28\linewidth]{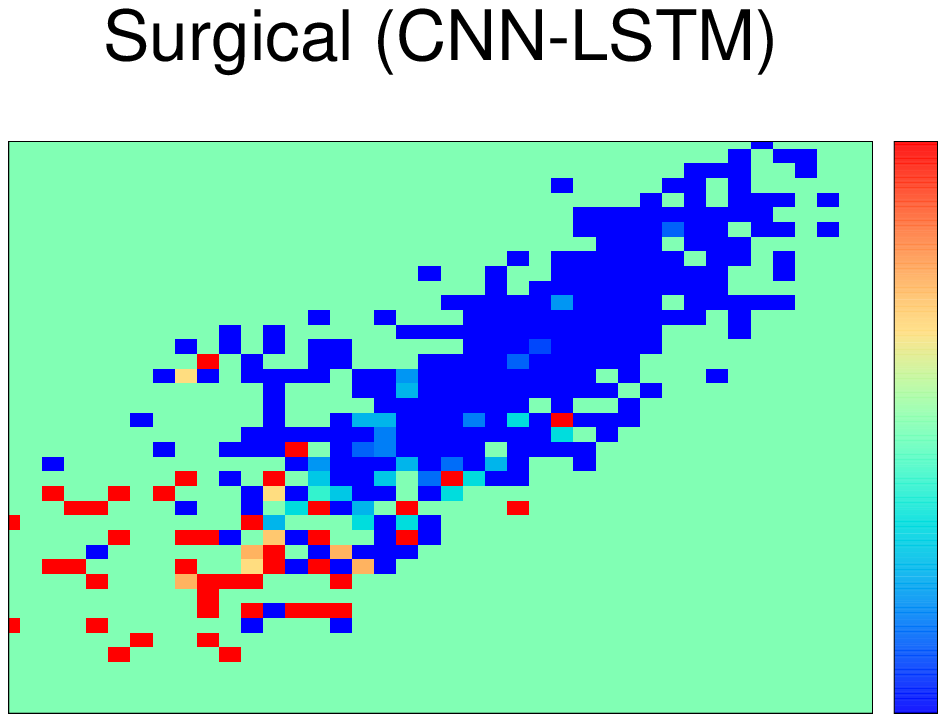}
	\end{center}
	\caption{(Color online) Mortality risk space for different ICU domains. Regions in red are risky. Each axis is a t-SNE~\cite{tsne} non-linear combination of: (top row) physiological parameters, or (bottom row) features extracted by CNN$-$LSTM.}
	\label{fig:space}
\end{figure*}

\begin{figure*}[h!]
	\begin{center}
		\includegraphics[type=eps,ext=.eps,read=.eps,width=0.275\linewidth]{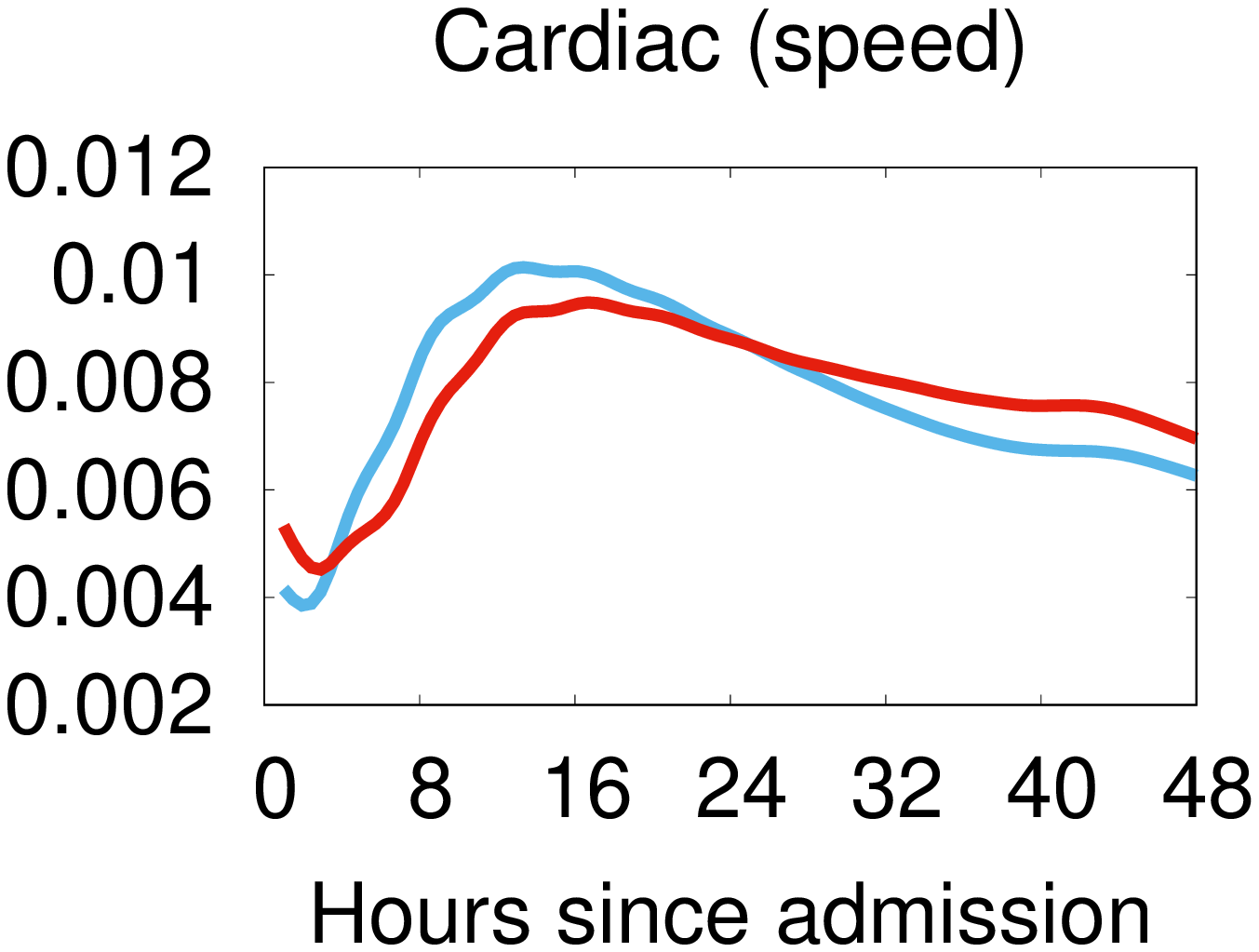}
		\hspace{-0.34in}
		\includegraphics[type=eps,ext=.eps,read=.eps,width=0.275\linewidth]{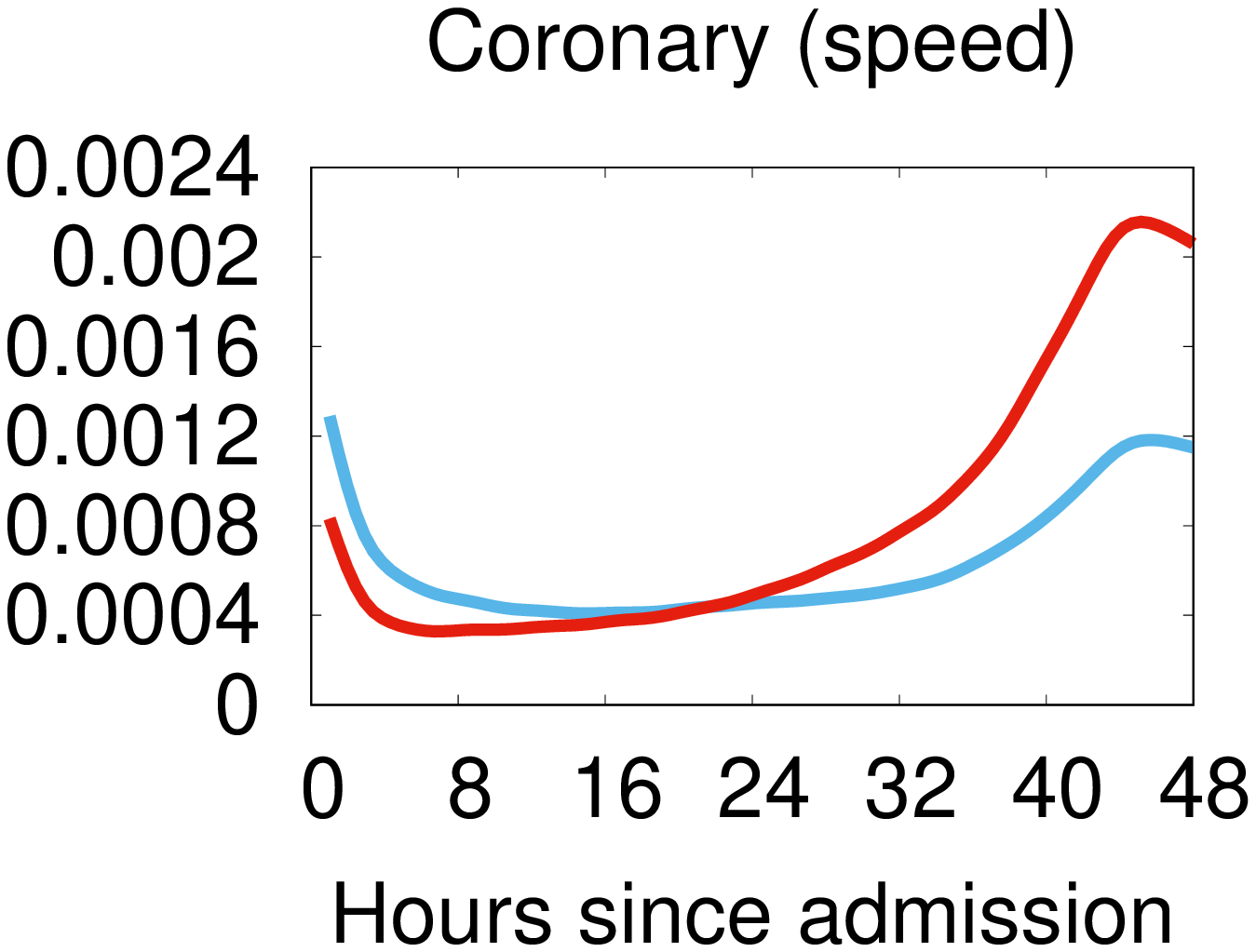}
		\hspace{-0.34in}
		\includegraphics[type=eps,ext=.eps,read=.eps,width=0.275\linewidth]{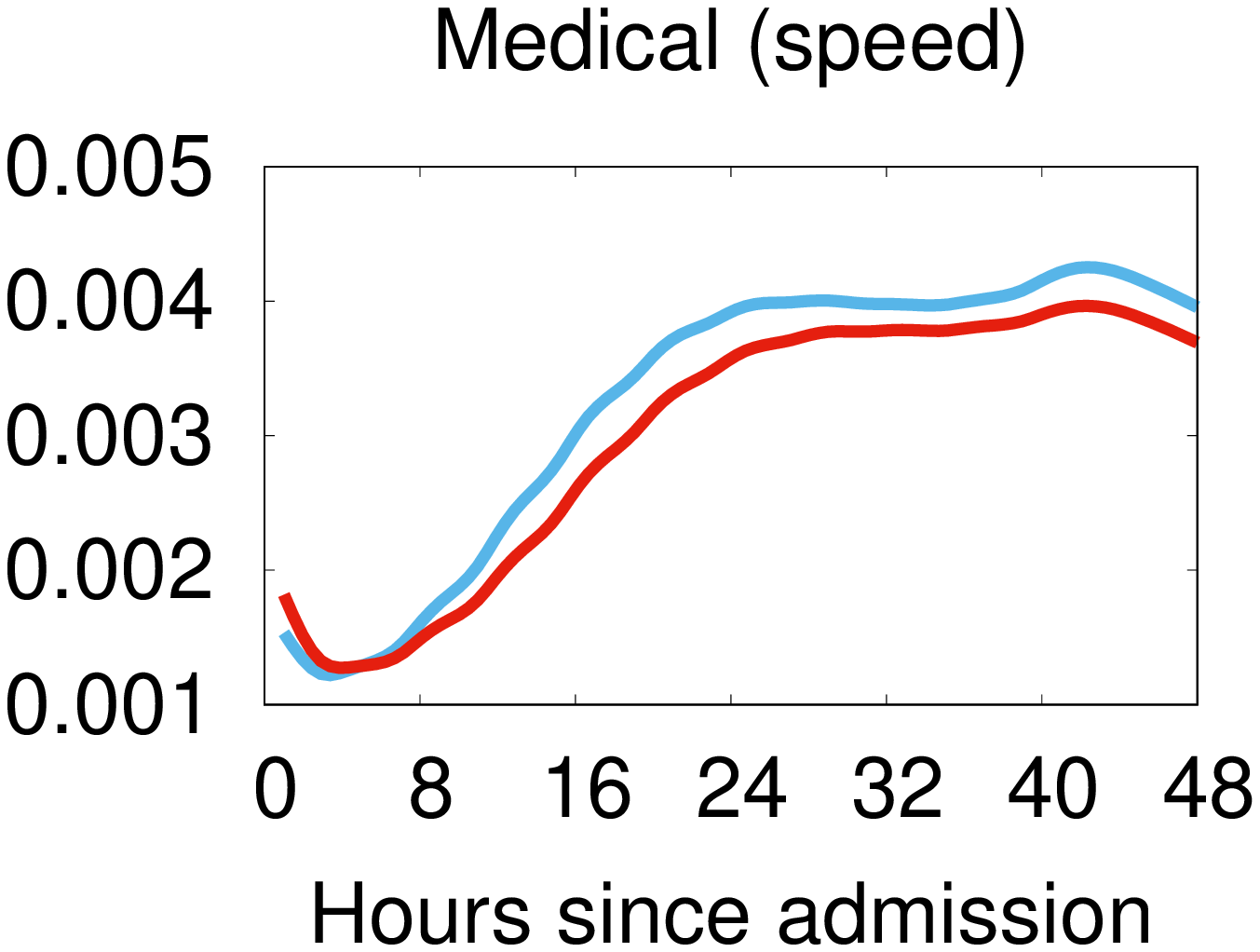}
		\hspace{-0.34in}
		\includegraphics[type=eps,ext=.eps,read=.eps,width=0.275\linewidth]{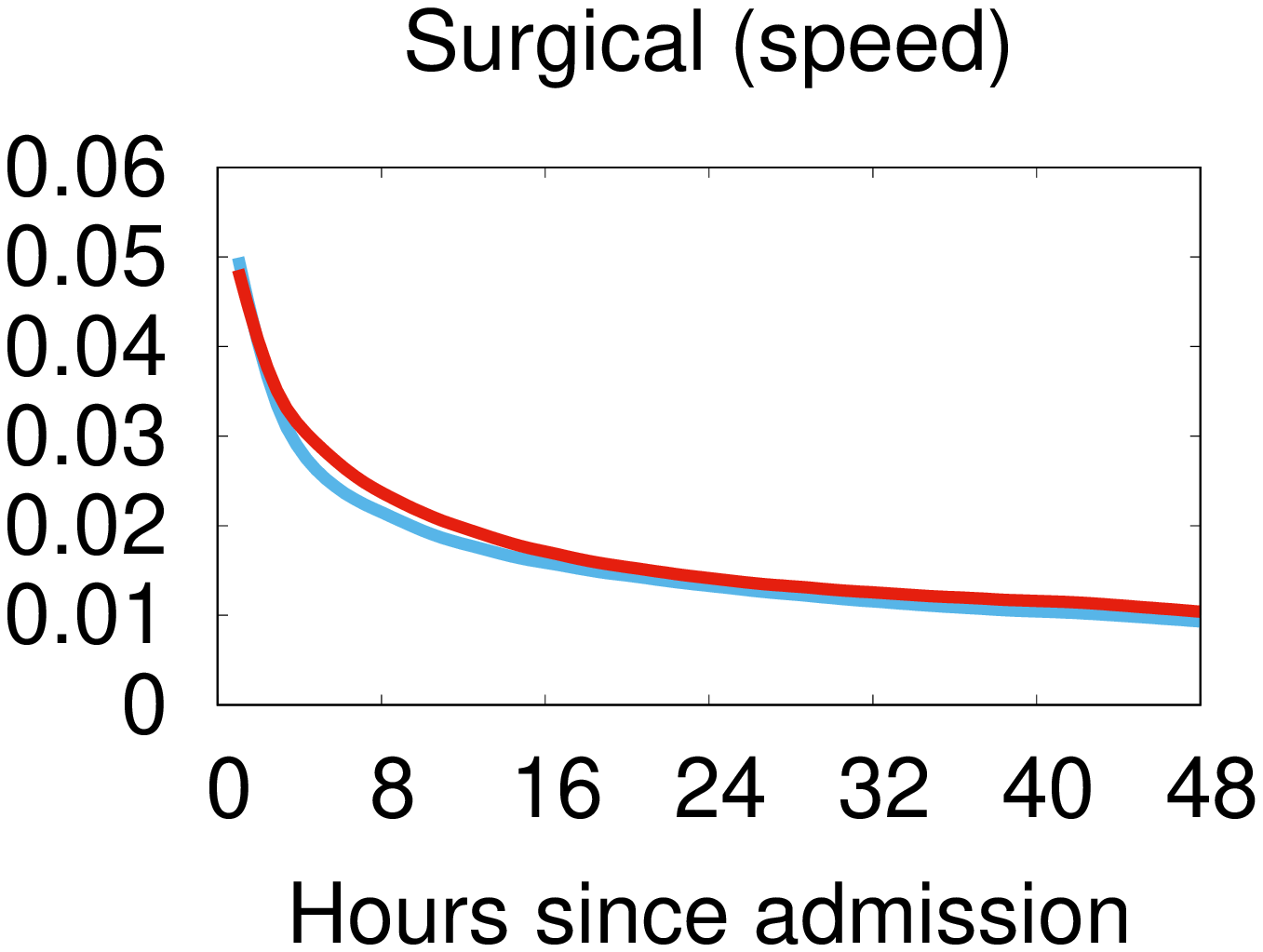}

		\includegraphics[type=eps,ext=.eps,read=.eps,width=0.275\linewidth]{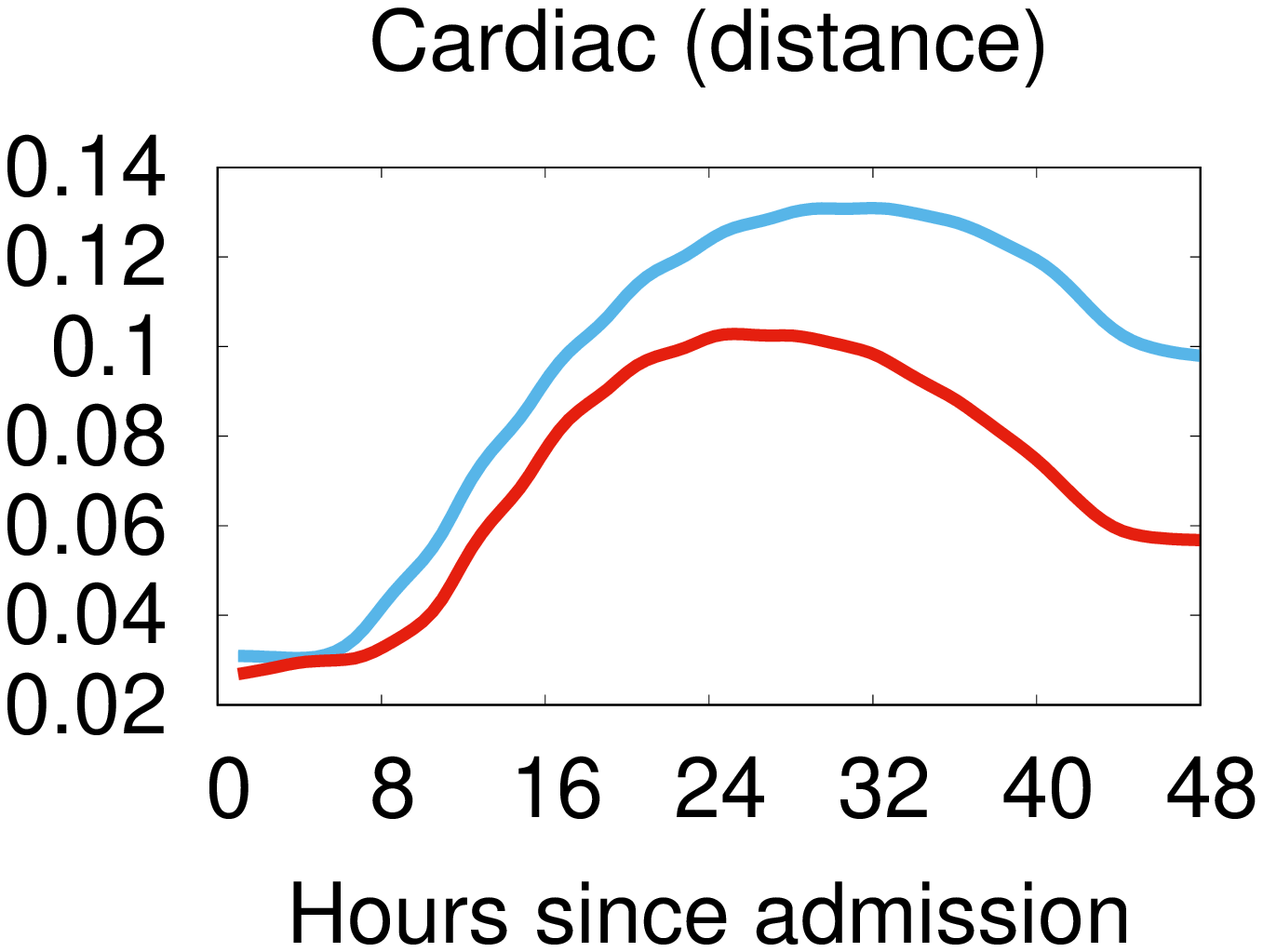}
		\hspace{-0.34in}
		\includegraphics[type=eps,ext=.eps,read=.eps,width=0.275\linewidth]{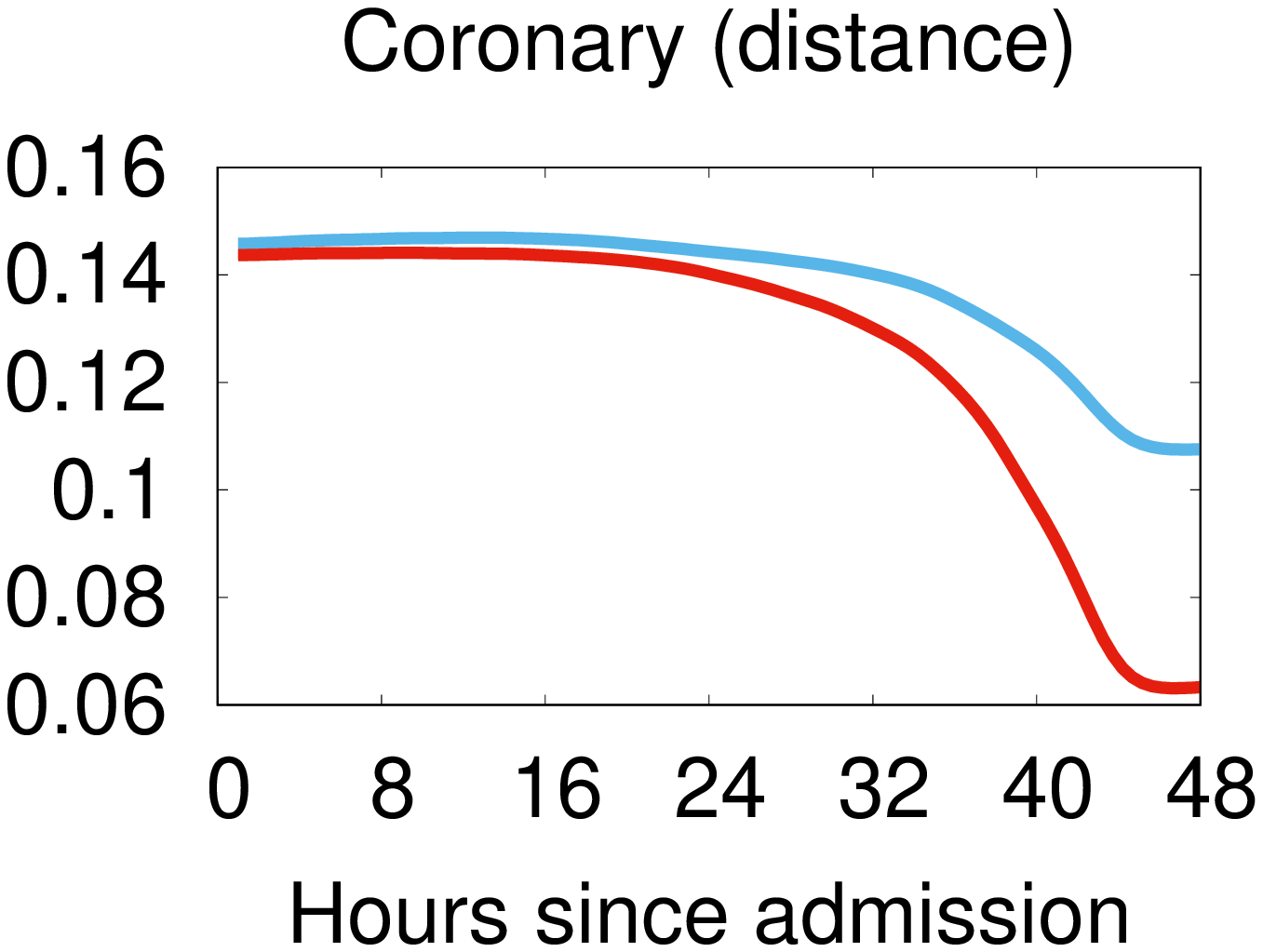}
		\hspace{-0.34in}
		\includegraphics[type=eps,ext=.eps,read=.eps,width=0.275\linewidth]{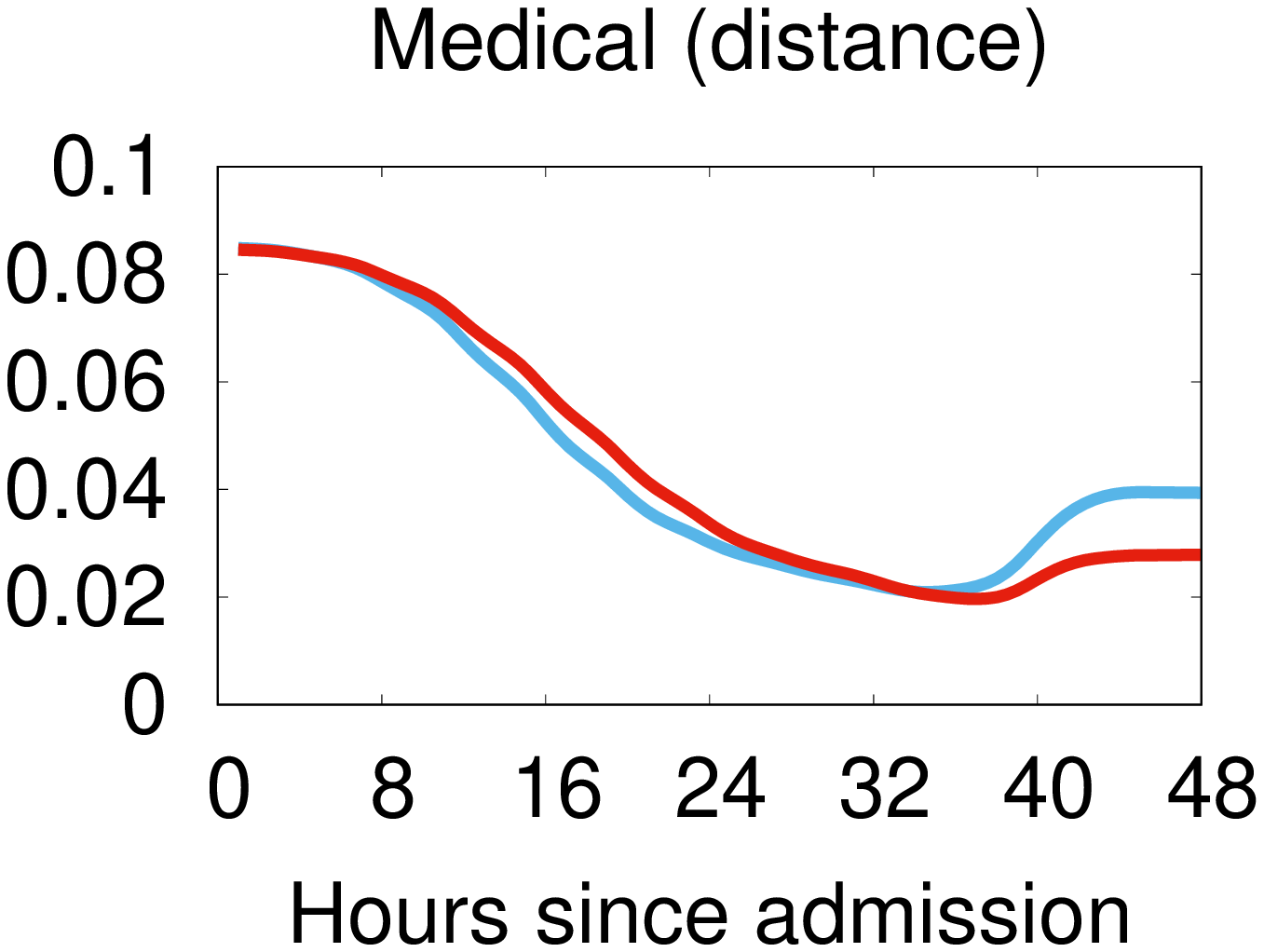}
		\hspace{-0.34in}
		\includegraphics[type=eps,ext=.eps,read=.eps,width=0.275\linewidth]{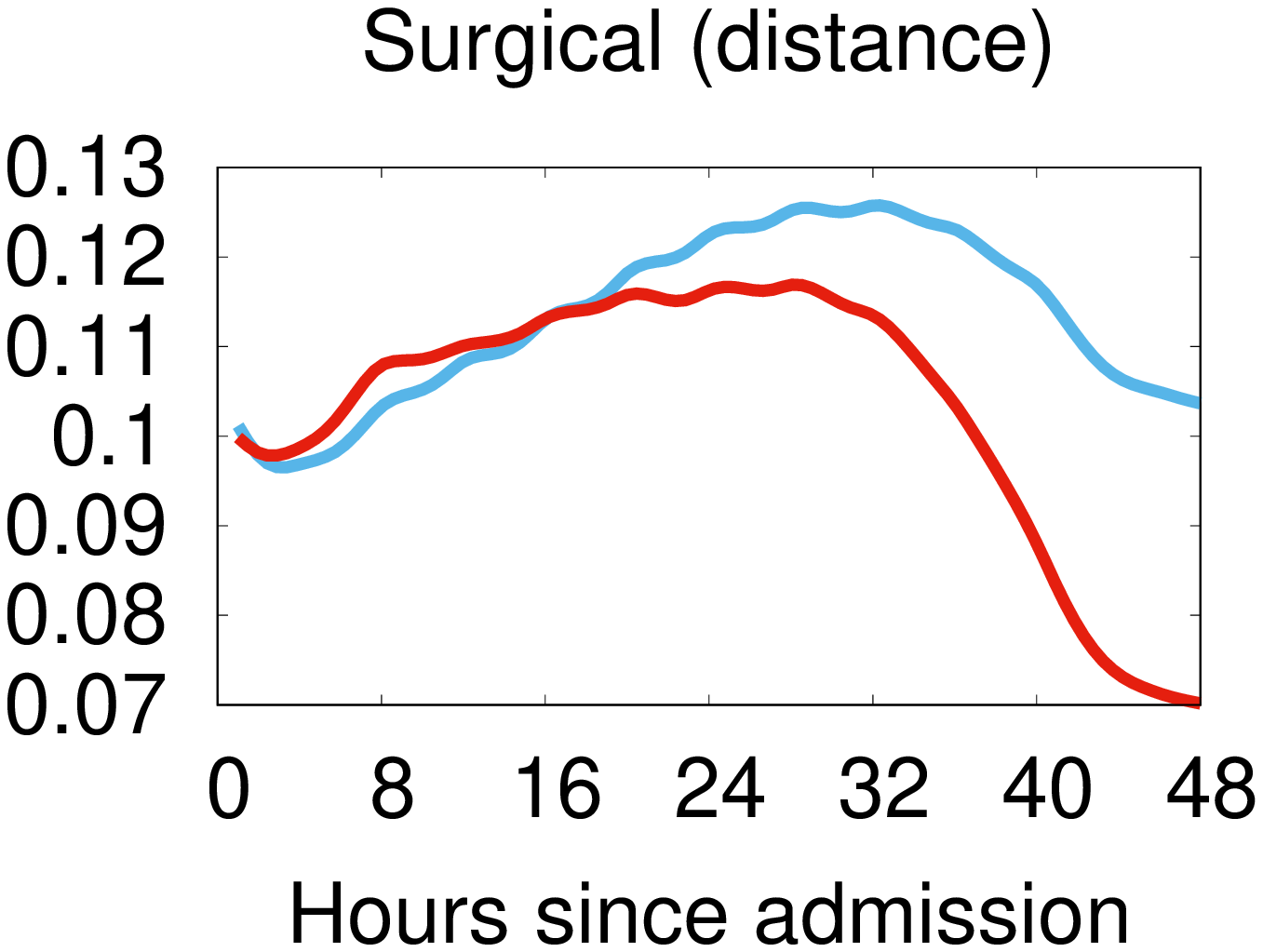}
	\end{center}
	\caption{(Color online) Dynamics of 48-hour trajectories in different ICU domains. Red curves are computed from trajectories associated with patients that have died. Blue curves are computed from trajectories associated with patients that survived.}
	\label{fig:dynamics}
\end{figure*}

The next set of experiments is devoted to answer Q3. Figure~\ref{fig:time1} shows AUC numbers obtained with predictions performed using information acquired within the first $x$ hours after the admission. AUC increases as more information is acquired. From the first 5 to 20 hours, the slopes associated with Cardiac and Coronary domains increase much faster than the slopes associated with Medical and Surgical domains. Figure~\ref{fig:time2} shows the gains obtained when compared with~\cite{kdd} at different prediction times. Early predictions performed by the CNN$-$LSTM architecture are much more accurate than those performed by~\cite{kdd}, particularly in the first hours after admission. The 10$-$20 hours period concentrates the more impressive gains, which vary from 4\% (Medical) to almost 8\% (Coronary).

The last set of experiments is concerned with Q4, i.e., to assess how meaningful are the mortality risk spaces.
Figure~\ref{fig:space} shows risk spaces for each ICU domain. These spaces are obtained by gathering patient trajectories, that is, the coordinates (i.e., CNN$-$LSTM representations) along with the predicted outcome at each time. Risk spaces can also be obtained from raw data and, in this case, the coordinates are simply the entire feature-vector. Risk spaces created from CNN$-$LSTM representations are much more meaningful than the corresponding spaces obtained from raw data.

Time is also encoded in the risk spaces, and thus we can exploit dynamics, such as the distance to the death centroid or the speed in which the patient condition changes. Figure~\ref{fig:dynamics} shows such dynamics in mortality risk spaces obtained from CNN$-$LSTM representations. Dynamics associated with the mortality risk space for the Cardiac and Coronary ICU domains are highly discriminative since red and blue curves are separated in the first hours after the patient admission. This may explain the high AUC numbers obtained in these domains. Patients show distinct dynamics, depending on the ICU domain, i.e., patients admitted to the Cardiac and Surgical units move much faster than patients admitted to the Coronary and Medical units. Also, the speed increases over time for patients admitted to the Coronary and Medical units.

\subsection{Feature Importance Estimates}

In order to interpret our model we chose to use a model agnostic representation of feature importance, where the impact of each feature on the model is represented using Shapley Additive Explanations, or simply SHAP~\cite{shap}. SHAP values provide a theoretically justified method for allocation of credit among a group. In our models, the group is a set of interpretable input feature values, and the credit is the prediction made by the model when given those input feature values. Specifically, feature importance is defined as the change in prediction probability when a feature is observed vs. unknown. Some feature values have a large impact on the prediction, while others have little impact. Unless otherwise stated, we used our CNN$-$LSTM model.

\begin{figure}[t]
	\begin{center}
                \hspace{-0.1in}
		\includegraphics[type=eps,ext=.eps,read=.eps,width=.94\linewidth]{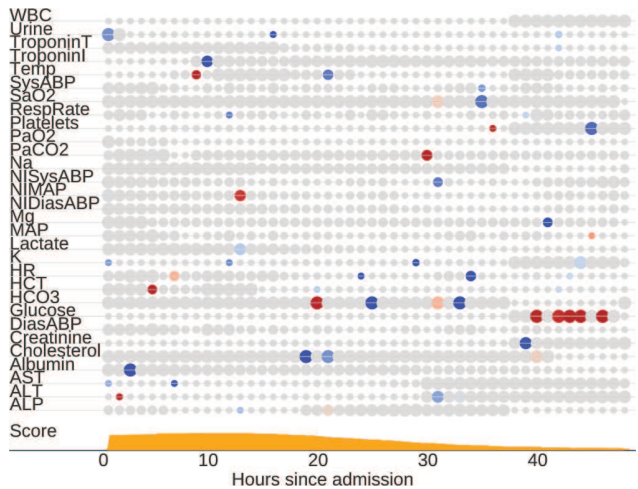}
                \hspace{0.1in}
	\end{center}
\caption{(Color online) A patient who has survived the hospitalization in the Cardiac ICU. Each row shows a physiological parameter and how its importance varies with time. Diameter is proportional to the parameter value. Color indicates SHAP values: red points push the mortality risk higher, while blue points push the mortality risk lower.}
	\label{fig:surv1}
\end{figure}

Figure~\ref{fig:surv1} shows a summary plot associated with a patient who has survived hospitalization. During the first hours of stay, the patient showed a mix of features contributing to survival and features contributing to death. The overall picture improved after the first 20 hours after admission, and the mortality risk has decreased significantly. Interesting to notice that our model was able to capture known but complex relationships, such as high glucose values inhibiting HCO3~\cite{hco3}.

\begin{figure}[t]
	\begin{center}
		\includegraphics[type=eps,ext=.eps,read=.eps,width=.94\linewidth]{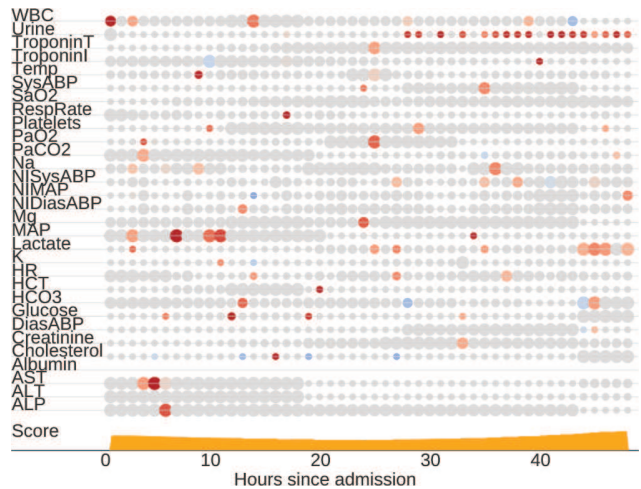}
	\end{center}
\caption{(Color online) A patient who has not survived the hospitalization in the Medical ICU. Each row shows a physiological parameter and how its importance varies with time. Diameter is proportional to the parameter value. Color indicates SHAP values: red points push the mortality risk higher, while blue points push the mortality risk lower.}
	\label{fig:surv2}
\end{figure}

Figure~\ref{fig:surv2} shows a summary plot associated with a patient who has not survived hospitalization. The patient showed a large number of features contributing to death. Medical interventions have stabilized some of the physiological parameters, but then other parameters started contributing to death. In particular, low urine output is often used as a marker of acute kidney injury~\cite{urine2} and long-term low urine output increases lactate levels~\cite{urine}. Despite changes in the physiological parameter values, mortality risk was always high for this patient.

\section{Conclusions}

ICU mortality prediction is a domain-specific problem. Thus, a prediction model learned from a sub-population of patients is likely to fail when tested against data from other population. We investigated this problem by considering four sub-populations of patients that were admitted to different ICU domains. We showed that patients within a specific ICU domain are physiologically different from patients within other domains. Nevertheless, patients across ICU domains still share basic characteristics. This motivates us to propose mortality prediction models based on domain adaptation. Specifically, our models learn domain invariant representations from time series ICU data while transferring the complex temporal dependencies between ICU sub-populations. The proposed models employ temporal feature extractors, being thus able to perform dynamic predictions during the ICU stay, potentially leading to earlier diagnosis. Finally, our models produce a mortality risk space, and the dynamics associated with patient trajectories are meaningful and can be very discriminative, enabling clinicians to track risky trends and to gain insight into their treatment decisions. Our models provide significant gains (4\% to 8\%) for early predictions, i.e., predictions within the first $5-20$ hour period after admission. Gains (2.5\% to 5\%) are also observed for predictions performed based on information acquired during the first 48 hours after admission.

	\vspace{0.2in}

\noindent\textbf{Acknowledgements.} We thank the partial support given by the project MASWEB - Models, Algorithms and Systems for the Web (grant FAPEMIG/PRONEX/MASWeb APQ-01400-14), and authors' individual grants and scholarships from CNPq, Fapemig and Kunumi. We gratefully acknowledge the support of NVIDIA Corporation with the donation of the Tesla K40 GPU used for this research.

	\bibliographystyle{IEEEtran}
	\bibliography{adriano}

\end{document}